\pdfoutput=1
\documentclass[runningheads]{llncs}

\usepackage[T1]{fontenc}
\usepackage{graphicx}
\usepackage{amsmath}
\usepackage{amssymb}
\usepackage{booktabs}
\usepackage[protrusion=true,expansion=false]{microtype}
\newcommand{\qwenmodel}{\texttt{Qwen2.5-Coder-}\allowbreak\texttt{1.5B-Instruct}}
\newcommand{\fullpassbest}{\texttt{full-pass@1\,=}\allowbreak\texttt{\,91.5\%}}
\newcommand{\maxnewtokens}{\texttt{max\_new\_tokens}\allowbreak\texttt{=768}}

\begin{document}

\title{Context-Instrumental Data Distillation for Kubernetes Manifest Generation: Method and Experimental Evaluation}

\titlerunning{Context-Instrumental Data Distillation for K8s Manifests}

\author{Andrey Kozachok\inst{1}\orcidID{0000-0001-6191-8614}\thanks{Corresponding author.} \and
Anatoliy Bakaev\inst{1}\orcidID{0000-0002-9526-0117} \and
Aleksandr Kozachok\inst{1}\orcidID{0000-0002-6501-2008} \and
Shamil Magomedov\inst{1}\orcidID{0000-0001-8560-1937} \and
Artem Noev\inst{1}\orcidID{0009-0003-2019-8536}}

\authorrunning{A. Kozachok et al.}

\institute{RTU MIREA, Moscow, Russia\\
\email{\{kozachok, bakaev, kozachok\_a, magomedov\_sh, noev\_a\}@mirea.ru}}

\maketitle

\begin{abstract}
This paper examines the specialization of Small Language Models~(SLMs) with up to 4~billion
parameters for generating artifacts in domain-specific languages~(DSL). Kubernetes manifests are
chosen as the target domain. We propose the \textbf{context-instrumental data distillation} method:
the source corpus is formed through synthetic generation and, in an extended scheme, through reverse
instruction generation from real Kubernetes YAML files, with pairs included in training only upon
passing external validators and matching the domain context model. Unlike classical KL-divergence
knowledge distillation, the baseline implementation reduces to supervised fine-tuning on
instrumentally verified examples. The experimental section presents a pilot implementation under
resource-constrained conditions: the DeepSeek-V4~Flash API serves as the teacher for synthetic
generation, while \qwenmodel{} is fine-tuned via LoRA on CPU. On the
\texttt{K8s-Distill-Pilot} corpus (\texttt{train\_1200}, \texttt{validation\_100},
\texttt{test\_200}), we achieved \fullpassbest{} (183/200) with a stricter prompt
formulation and \maxnewtokens{}. The key empirical finding is that for Kubernetes
YAML, result quality in the pilot depended more on strict output format requirements than on simply
increasing the number of training examples.

\end{abstract}

%
% ============================================================
\section{Introduction}
% ============================================================

\subsection{Motivation and Problem Statement}

IT infrastructure management increasingly relies on the IaC (Infrastructure as Code) paradigm,
where resource state is described through declarative specifications rather than manual
administrator actions. In this paradigm, Kubernetes has become one of the primary container
orchestration platforms. Manifest complexity has grown significantly: a typical production
application configuration may include dozens of interconnected objects (Deployment, Service,
Ingress, NetworkPolicy, HPA, RBAC), and an error in any one of them can cause an incident.

Industry reports on cloud infrastructure reliability and security indicate that configuration
errors and insufficient change control remain notable sources of operational risk~\cite{ref1,ref2}.
Practitioners also deal with a growing set of DSLs and infrastructure formats. General-purpose
Large Language Models~(LLMs), including GPT-4 and Llama-3-70B-Instruct, can assist in such tasks,
but their use in infrastructure scenarios is limited for several reasons:
\begin{enumerate}
  \item Transmitting configurations of critical information infrastructure to external cloud APIs
        may be prohibited under regulatory requirements and internal organizational policies.
  \item Running LLMs for inference requires substantial computational resources, making
        large-scale configuration generation expensive.
  \item LLMs may make errors in DSL parameters, e.g., generating incorrect API groups or
        violating resource schemas.
\end{enumerate}

\subsection{Small Language Models and Specialization}

One practical solution is the use of Small Language Models (SLMs, approximately 1--4 billion
parameters), which can be run locally or in an isolated environment on limited resources. Work from
2024 on \textbf{Qwen2.5-Coder}~\cite{ref3} and \textbf{Phi-3/Phi-3.5}~\cite{ref4} shows that
with high-quality training data, compact models can compete with larger ones in code generation and
instruction-following tasks.

However, standard fine-tuning methods on open-source code corpora do not account for DSL
specifics: high density of formal rules and low error tolerance. Effective SLM specialization
requires a mechanism that not only imitates the style of YAML files but also accounts for the
strict grammatical and semantic constraints of the domain.

This paper proposes the \textbf{context-instrumental data distillation} method. It leverages the
formal nature of Kubernetes to build a verified training set and subsequently specialize a compact
student model on correct domain examples. The full method includes \texttt{synthetic\_direct} and
\texttt{real\_reverse} streams. In the conducted pilot, the main corpus was formed through the
\texttt{synthetic\_direct} stream and targeted error-correction batches.

Throughout this paper, ``distillation'' is used in the sense of \emph{data distillation}, i.e.,
selecting and transferring verified source examples from a teacher model—and in the extended scheme
also from real sources—to the student model. The method does not assume preservation of the teacher
model's unnormalized output scores (logits), and in the baseline implementation is not
KL-divergence distribution distillation.

%
% ============================================================
\section{Related Work}
% ============================================================

Recent work on small models for programming has shifted focus from simply increasing corpus size to
data filtering quality, synthetic generation, and reproducibility of data preparation.

\subsection{Advances in SLMs for Code}

The \textbf{Qwen2.5-Coder} family (September~2024) was trained on a large code corpus with
enhanced cleaning, synthetic generation, and data balancing~\cite{ref3}. This aligns with the
thesis of~\cite{ref5} that for code models the density of useful knowledge in the training set
matters more than the number of parameters.

However, in DSL tasks even strong code models retain the risk of structural errors, as they are
trained on large collections of general-purpose languages where acceptable syntactic variability is
higher than in configuration schemas.

\subsection{Distillation and Data Synthesis Methods}

Research on knowledge distillation shows a clear shift toward using Chain-of-Thought~(CoT)
reasoning and reproducible data synthesis procedures. \textbf{OSS-Instruct}~(Magicoder~\cite{ref6})
and \textbf{Evol-Instruct}~(WizardCoder~\cite{ref7}) demonstrated that training pairs can be
obtained from open-source code fragments. More recent works, including DataDreamer~\cite{ref8} and
Source2Synth~\cite{ref9}, separately emphasize the role of reproducibility, anchoring generation
to source data, and filtering low-quality examples.

For Kubernetes and other IaC tools, such procedures are typically insufficient. Evaluation via an
LLM judge or general tests poorly reflects the schema constraints and security policies of
infrastructure DSLs. In the proposed approach, such evaluation is replaced by \textbf{instrumental
verification}, where an example enters the distillation corpus only after passing syntactic,
schema, semantic, and security checks.

A methodologically close line of work was developed at ISP RAS in verification of system
software. In the Linux Driver Verification project, rules for correct kernel API usage are
formalized, grouped by category, and expressed as machine-readable templates, then checked by an
integrated static verification platform~\cite{ref10}. Works on AstraVer and formal security policy
models also demonstrate the practical role of contract specifications, SMT solvers, and
machine-verifiable security models~\cite{ref11,ref12}. Unlike these directions, the present paper
uses instrumental verification not to prove the correctness of existing code, but to filter the
source distillation corpus before SLM fine-tuning.

\subsection{Grammar-Constrained Decoding}

Works by \textbf{Willard \& Louf}~\cite{ref13} and Geng et al.~\cite{ref14} describe
grammar-constrained decoding~(GCD), where invalid tokens are excluded at inference time. This
approach has a known limitation: strict grammatical decoding can distort the model's natural
distribution, so it is beneficial to combine inference constraints with training on correct domain
examples for structured generation~\cite{ref15}. In this paper, GCD is considered an auxiliary
format-control mechanism. The primary transfer of domain behavior is achieved through training on
instrumentally verified examples.

%
% ============================================================
\section{Theoretical Model of Context-Instrumental Data Distillation}
% ============================================================

In narrow formal domains, a model must reproduce the style of examples while remaining within the
allowable space of schemas and policies. Therefore, domain behavior transfer is defined through a
context model and a set of instrumental checks.

\subsection{Formalization of the DSL Context Model}

By \emph{context model}~(CM) we mean a structured domain description that serves as an inductive
bias for the model. For Kubernetes, the CM is defined as an ordered tuple:
\begin{equation}
  CM_{K8s} = \langle \mathcal{M},\, \mathcal{G},\, \mathcal{S},\, \mathcal{P},\, \mathcal{V} \rangle
\end{equation}
where:
\begin{itemize}
  \item $\mathcal{M}$ (meta-specification): a set of tuples $\langle \mathit{version},\,
        \mathit{kind} \rangle$ defining allowed resource types.
  \item $\mathcal{G}$ (grammar): a subset of context-free grammar $G=(V,\Sigma,R,S)$ describing
        YAML syntax projected onto Kubernetes specifications.
  \item $\mathcal{S}$ (schema/API): a formal description of resource attributes based on the
        OpenAPI schema, including data types, field requirements, and constraints.
  \item $\mathcal{P}$ (typical compositions): a knowledge base of typical artifact compositions
        (e.g., the \texttt{Deployment}\,$\to$\,\texttt{Service}\,$\to$\,\texttt{Ingress} chain).
  \item $\mathcal{V}$ (validators): a set of mappings $v\colon A \to \{0,1\}$, where $A$ is the
        space of generated artifacts, returning~1 when formal correctness conditions are met.
\end{itemize}

\subsection{Mathematical Formulation of the Specialization Task}

In classical knowledge distillation~\cite{ref16}, the Kullback--Leibler divergence between teacher
token distribution $P_T$ and student distribution $P_S$ is minimized. In our baseline
implementation, teacher model logits are not preserved, so the task is formulated not as
KL-divergence distribution distillation but as supervised training on an instrumentally filtered
source corpus.

Let the source corpus be:
\begin{equation}
  \mathcal{D}_C = \{(x_i,\, c_i,\, y_i,\, s_i)\}_{i=1}^{N},
\end{equation}
where $x_i$ is a natural-language query, $c_i$ is a fragment of the context model, $y_i$ is a
Kubernetes artifact, and $s_i \in \{\texttt{synthetic\_direct},\, \texttt{real\_reverse}\}$ is the
source of the pair. For \texttt{synthetic\_direct}, the artifact is generated by the teacher; for
\texttt{real\_reverse}, the artifact is taken from a real YAML corpus, and the teacher generates
only the instruction. In the conducted pilot, the final training split was built primarily from
\texttt{synthetic\_direct} and targeted error-correction batches; therefore, \texttt{real\_reverse}
is considered a method extension. The instrumental circuit defines a binary filter:
\begin{equation}
  \mathbb{I}_{\mathcal{V}}(y_i) =
  \begin{cases}
    1, & \text{if } y_i \text{ passes all validators } \mathcal{V},\\
    0, & \text{otherwise}.
  \end{cases}
\end{equation}
After filtering, the corpus is formed:
\begin{equation}
  \mathcal{D}_{\mathcal{V}} =
  \{(x_i, c_i, y_i, s_i) \in \mathcal{D}_C \mid \mathbb{I}_{\mathcal{V}}(y_i)=1\}.
\end{equation}
The student model $P_S$ is trained by minimizing the negative log-likelihood:
\begin{equation}
  \mathcal{L}_{CI} =
  -\,\mathbb{E}_{(x,c,y)\sim\mathcal{D}_{\mathcal{V}}}
  \!\left[
    \sum_{t=1}^{|y|} \log P_S\!\left(y_t \mid y_{<t}, x, c\right)
  \right].
\end{equation}

The instrumental filter does not change the loss function at the token level but changes the
distribution of training examples. The student receives gradients only from trajectories that
satisfy the schema, semantic constraints, and domain security policies. In an extended
implementation that preserves soft target distributions or teacher logits, the method can be
supplemented with a classical KL term; this variant is not used here.

Corpus representativeness in the full scheme is monitored separately from validity. For corpus
$\mathcal{D}$ and reference slice $\mathcal{R}$, a feature vector $q(\mathcal{D})$ is constructed:
distribution of GVK, resource families, co-occurrence of resources, and complexity levels. The
divergence between $q(\mathcal{D}_{\mathcal{V}})$ and $q(\mathcal{R})$ is measured via
Jensen--Shannon divergence, total variation distance, and rare class coverage. In the pilot, this
part is fixed as a limitation and a direction for extension: instrumental validity does not
guarantee alignment with the industrial distribution of Kubernetes artifacts.

\subsection{Domain Shift Effect}

In this paper, \emph{domain shift} refers to constraining the model's generative behavior within
the boundaries set by $CM$ and corpus $\mathcal{D}_{\mathcal{V}}$; a related concept is sometimes
described in the literature as ``epistemic isolation.'' This is not about the absence of external
knowledge—the base SLM is already pre-trained on a broad code and text corpus, retaining both
general knowledge and associations irrelevant to Kubernetes. Fine-tuning on instrumentally verified
domain examples shifts the probability mass toward validated Kubernetes templates and reduces the
influence of irrelevant skills not related to the target DSL.

%
% ============================================================
\section{Implementation of the Context-Instrumental Data Distillation Process}
% ============================================================

\begin{figure}[htbp]
\centering
\includegraphics[width=\textwidth]{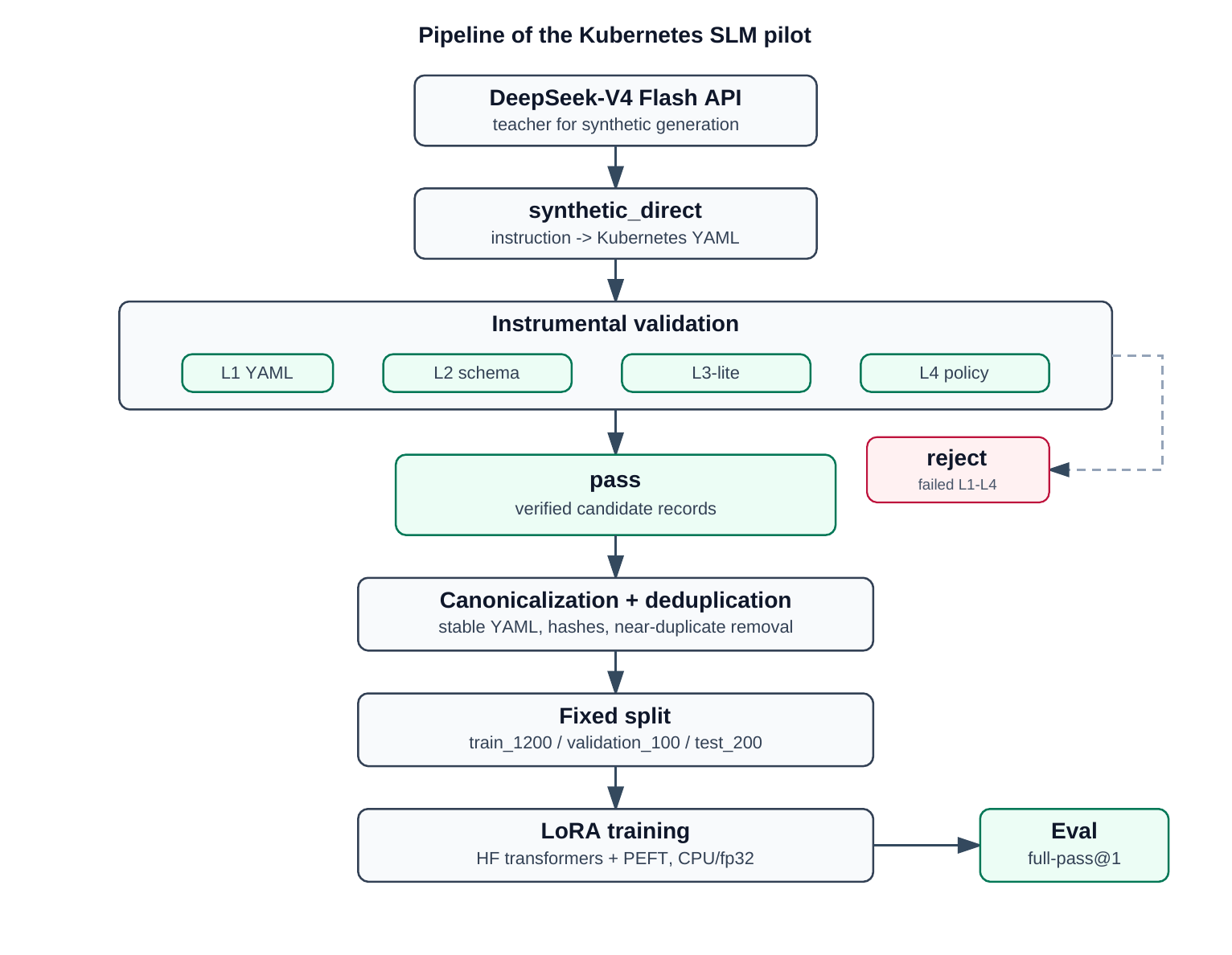}
\caption{Pilot experiment pipeline: generation via API, L1--L4 filtering, canonicalization,
deduplication, fixed split, LoRA training, and evaluation on \texttt{test\_200}.}
\label{fig1}
\end{figure}

Behavior transfer from teacher to student in the IaC domain is divided into four technological
stages whose purpose is to obtain a corpus where examples reproduce the Kubernetes manifest format
and pass domain checks. The overall pilot pipeline structure is shown in Fig.~\ref{fig1}.

\subsection{Stage 1: Assembly of Source Pairs}

In the pilot implementation, the teacher model is the DeepSeek-V4~Flash API
(\texttt{deepseek-v4-flash}). The primary stream used was \texttt{synthetic\_direct}: the teacher
receives a structured prompt including the target resource family, security constraints, Kubernetes
schema version, and a requirement to return only YAML without Markdown blocks or explanations.

The extended method also provides a \texttt{real\_reverse} stream: a source Kubernetes YAML is
taken from a public source, normalized and cleaned, after which the teacher generates a
natural-language instruction that could have produced this YAML. Potential sources for such an
extension include public YAML and Kubernetes slices, including The Stack,
\texttt{substratusai/the-stack-yaml-k8s}, and Artifact Hub~\cite{ref17,ref18,ref19}. In the pilot,
\texttt{real\_reverse} did not become the primary training corpus source and was not used in the
best result. It remains a potential pipeline extension for testing external validity on real YAML
artifacts.

To increase synthetic stream diversity, stratification was applied across Kubernetes resource types
and residual schema error classes: RBAC, StatefulSet, CronJob, Ingress, NetworkPolicy, HPA,
ConfigMap/Secret, and composite YAML packages. In later stages, targeted correction tasks were
added to address residual schema and output format errors.

\subsection{Stage 2: Instrumental Cleaning}

This stage implements the filter function $\mathbb{I}_{\mathcal{V}}$. Each artifact passes a
multi-level verification circuit:
\begin{enumerate}
  \item \textbf{Syntactic level~(L1)}: validation of YAML structure. Artifacts with indentation
        violations, incorrect escaping, or natural language outside YAML are rejected.
  \item \textbf{Schema level~(L2)}: \texttt{kubeconform --strict}~\cite{ref20} with a local
        schema cache for Kubernetes~1.30.0. Checks include required fields, data type correctness,
        API group validity, and absence of unknown fields.
  \item \textbf{Semantic level~(L3-lite)}: limited cross-resource validation. The current
        implementation applies rules including \texttt{Service.spec.selector} matching Pod template
        labels and presence of \texttt{HPA.spec.scaleTargetRef} target in the YAML package.
        Therefore, the paper uses the notation \texttt{L3-lite} rather than claiming full
        Kubernetes semantic validation.
  \item \textbf{Security policies~(L4)}: \texttt{Checkov}~\cite{ref21} and
        \texttt{Trivy}~\cite{ref22} with a minimal set of critical policies. Other detected issues
        are saved as warnings and do not automatically discard records.
\end{enumerate}

\begin{figure}[htbp]
\centering
\includegraphics[width=\textwidth]{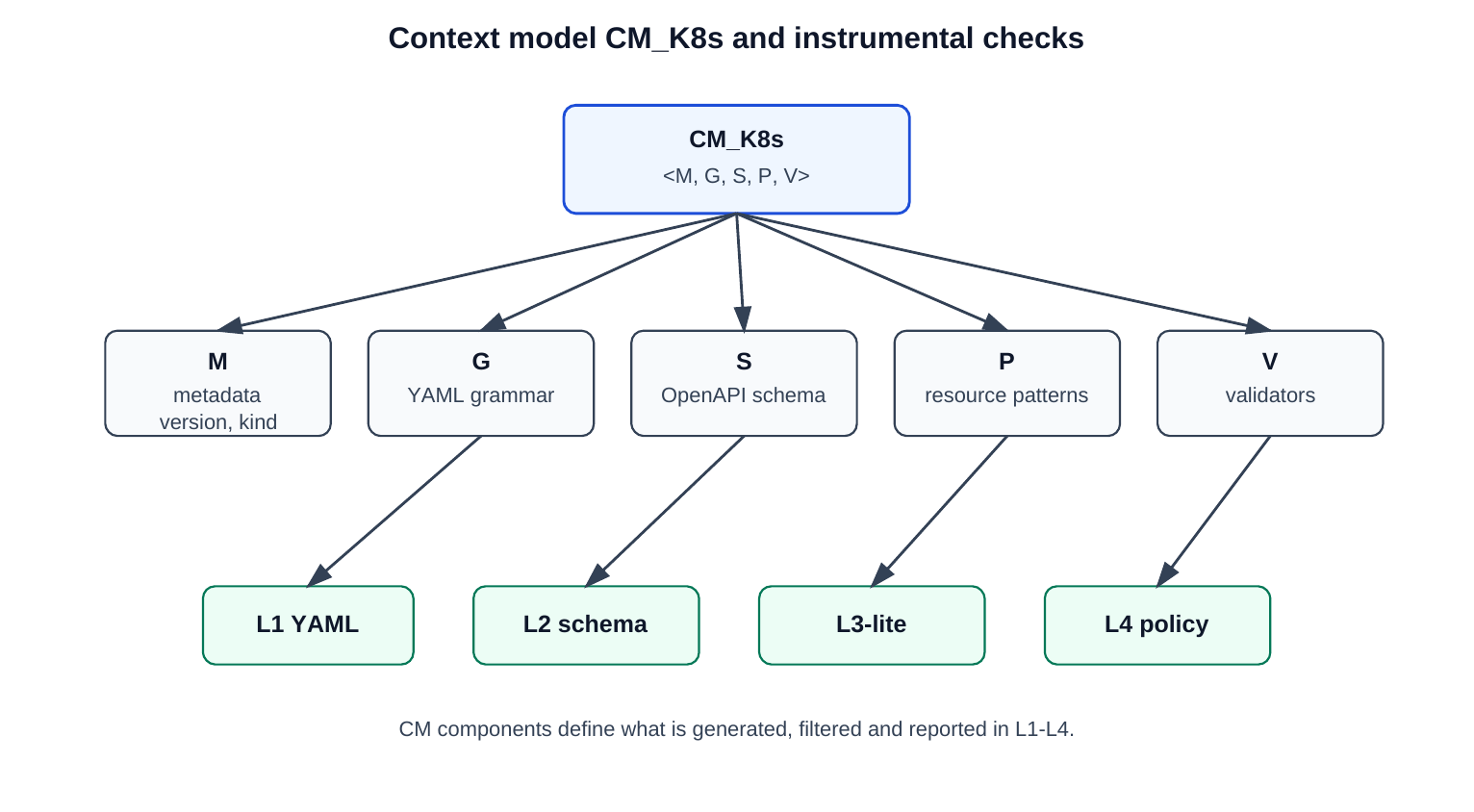}
\caption{Kubernetes context model components and their relationship to instrumental verification
levels L1--L4.}
\label{fig2}
\end{figure}

The relationship between context model components and instrumental verification levels is shown in
Fig.~\ref{fig2}.

Only pairs that pass L1--L4 in the specified configuration are included in the final pilot corpus.
Server-side validation via \texttt{kubectl --dry-run=server} was not performed in the pilot and is
considered a direction for future validation.

\subsection{Stage 3: Pilot Student Model Fine-Tuning}

In the pilot implementation, the primary student model is \qwenmodel{}.
Fine-tuning was performed as LoRA using Hugging Face Transformers and PEFT in CPU/fp32 mode. This is
a pilot LoRA run on CPU, not an industrial SFT pipeline via \texttt{trl.SFTTrainer}.

The actual configuration of the best run:
\begin{itemize}
  \item \textbf{Model}: \qwenmodel{};
  \item \textbf{Method}: LoRA, fp32;
  \item \textbf{LoRA rank}: $r = 4$;
  \item \textbf{LoRA alpha}: $\alpha = 8$;
  \item \textbf{Target modules}: \texttt{q\_proj}, \texttt{k\_proj}, \texttt{v\_proj},
        \texttt{o\_proj}, \texttt{gate\_proj}, \texttt{up\_proj}, \texttt{down\_proj};
  \item \textbf{Optimizer}: AdamW;
  \item \textbf{Mini-batch size}: 1 with gradient accumulation;
  \item \textbf{Hardware platform}: CPU, laptop with 32~GB RAM;
  \item \textbf{Evaluation mode}: Hugging Face Transformers and PEFT on CPU.
\end{itemize}

QLoRA~\cite{ref23}, NF4, and \texttt{Unsloth} were not used in the conducted pilot. They can only
be considered as options for a future full method implementation on GPU.

\subsection{Stage 4: Feedback Loop}

The fine-tuned model can be used to generate new examples. Artifacts that pass the verifier
$\mathcal{V}$ but show low quality on individual L1--L4 levels can become sources of targeted
correction tasks for the next iteration.

In the pilot, this feedback loop was applied in a limited form: after error analysis, targeted
replenishment and residual correction batches were formed. A full self-improvement cycle was not
performed.

%
% ============================================================
\section{Experimental Evaluation}
% ============================================================

\subsection{Pilot Corpus Characteristics}

In the pilot experiment, the \textbf{K8s-Distill-Pilot} corpus was formed from 1\,710 unique
verified records after canonicalization, L1--L4 filtering, and deduplication.

The actual split: \texttt{train\_1200} / \texttt{validation\_100} / \texttt{test\_200}.

The primary source of pairs was \texttt{synthetic\_direct} via the DeepSeek-V4 Flash API,
supplemented by targeted replenishment and residual correction batches.

The \texttt{test\_200} set was fixed before the main comparison runs and did not change between
iterations. Since it was formed from the same synthetic distribution as the training corpus,
results should be interpreted as an assessment of \textbf{in-domain generalization}, not as
evidence of transfer to arbitrary real-world IaC scenarios.

\subsection{Evaluation Metrics}

The pilot's primary metric is \texttt{full-pass@1}: an adaptation of the \texttt{pass@k}
metric~\cite{ref24} for instrumental evaluation, i.e., the fraction of responses passing
L1\,+\,L2\,+\,L3-lite\,+\,L4 with a minimal set of critical failures.

The pilot was evaluated using the following metrics:
\begin{enumerate}
  \item \textbf{SC}: syntactically valid YAML.
  \item \textbf{schema-pass@1}: passing L1--L2, including \texttt{kubeconform --strict}.
  \item \textbf{semantic-pass@1}: passing L1--L3-lite within the implemented rules.
  \item \textbf{policy-pass@1}: passing L1--L2 and the minimal set of critical L4 failures.
  \item \textbf{full-pass@1}: passing L1--L4 in the pilot configuration.
  \item \textbf{structural exact match}: match after YAML parsing and canonicalization.
  \item \textbf{BLEU\_auxiliary}: an auxiliary text metric, not used as the primary YAML quality
        criterion.
  \item \textbf{Resource metrics}: training time, evaluation time, peak RAM, and time per token
        in CPU mode.
\end{enumerate}

\texttt{full-pass@5}, multiple stochastic generations, 95\% confidence intervals over multiple
seeds, and server-side validation via \texttt{kubectl --dry-run=server} were not performed in this
pilot. These checks remain extensions of the experimental protocol.

\subsection{Results and Comparative Analysis}

\begin{figure}[htbp]
\centering
\includegraphics[width=\textwidth]{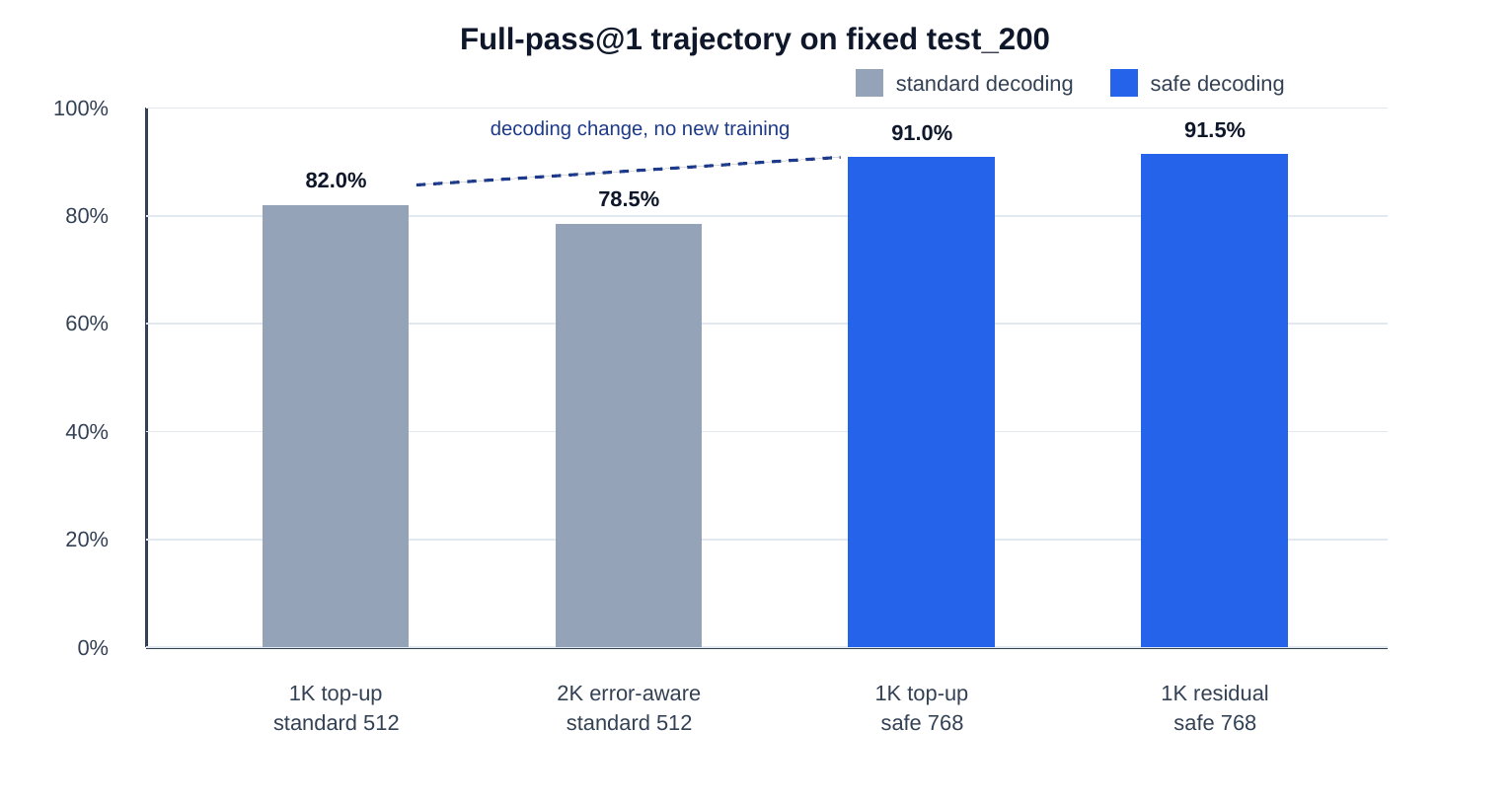}
\caption{\texttt{full-pass@1} trajectory on fixed \texttt{test\_200}. The main jump
$82.0\% \to 91.0\%$ was obtained by changing the inference mode, without retraining the adapter.}
\label{fig3}
\end{figure}

\begin{figure}[htbp]
\centering
\includegraphics[width=\textwidth]{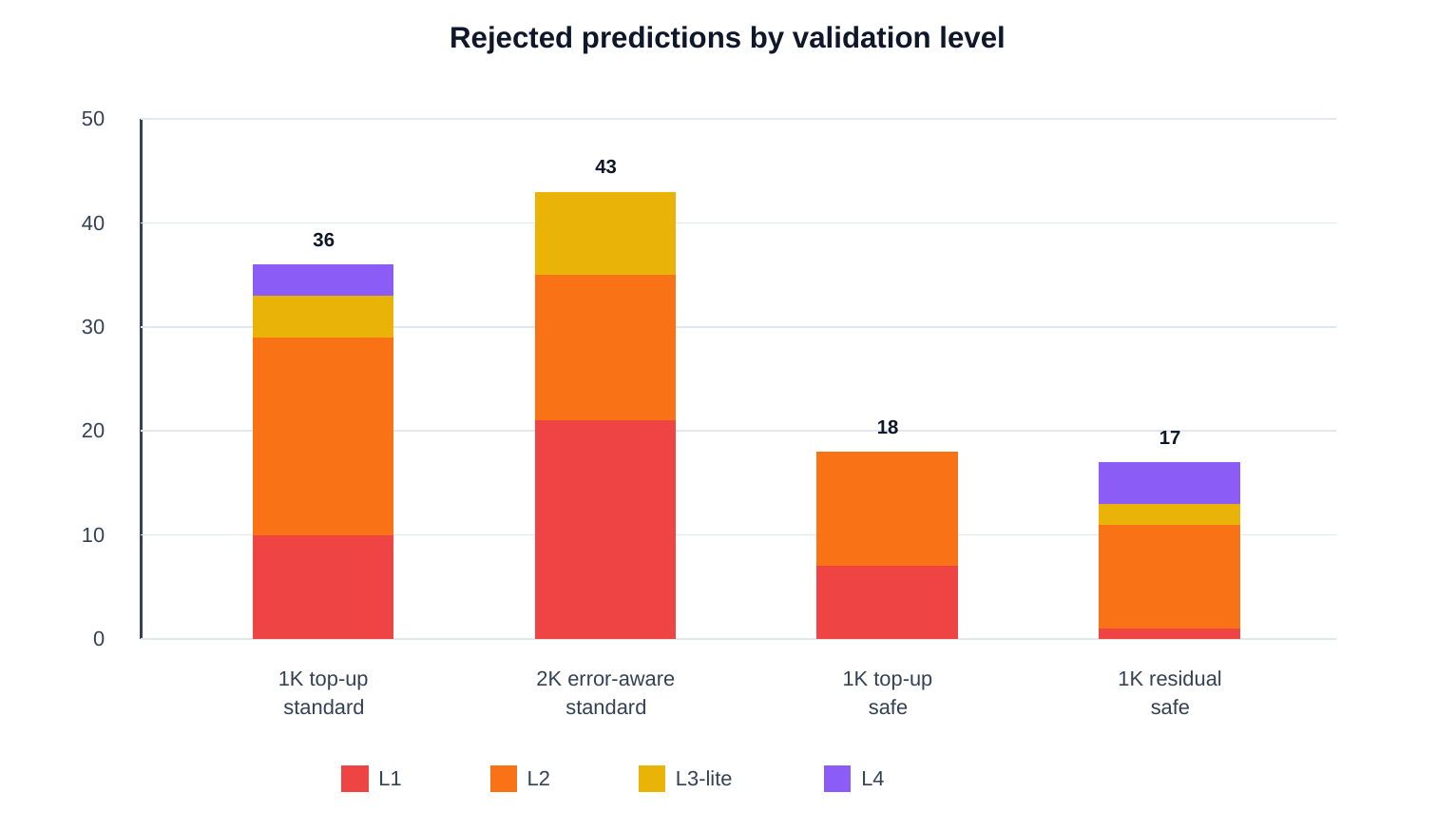}
\caption{Distribution of failures by L1--L4 levels for four comparable runs. In the best mode,
residual errors concentrate primarily at~L2.}
\label{fig4}
\end{figure}

The pilot compares not different models but the improvement trajectory of a single SLM baseline:
\qwenmodel{}~+~LoRA. The result shows that in Kubernetes YAML generation
the inference mode can matter more than simply increasing the training corpus. Table~\ref{tab1}
provides a numerical summary; Figs.~\ref{fig3} and~\ref{fig4} visualize the quality dynamics and
failure structure.

Table~\ref{tab1} compares four sequential pilot runs.
\textit{1K~+~diversity} uses a baseline training set of 1\,000 records supplemented with examples
for increasing diversity of names, namespaces, images, ports, selectors, and typical resource
combinations.
\textit{2K~+~error~corr.} increases the training set to 2\,000 records and adds examples targeted
at previously observed L1--L4 errors.
\textit{1K~+~strict~infer.} uses the same 1K adapter but changes the inference mode: a stricter
prompt formulation, prohibition of explanations outside YAML, and increased
\maxnewtokens{}.
\textit{1.2K~+~resid.~corr.} further trains the model on 200 targeted examples addressing residual
L2--L4 errors and is evaluated in the same strict inference mode.

\begin{table}[htbp]
\caption{Pilot run trajectory on the fixed \texttt{test\_200}.}
\label{tab1}
\centering
\scriptsize
\setlength{\tabcolsep}{3pt}
\begin{tabular}{@{}lrlrrrrrr@{}}
\toprule
Run & Train & Mode & full-pass & L1 & L2 & L3 & L4 & BLEU\\
\midrule
1K + diversity    & 1\,000 & std, 512    & 164/200 = 82.0\% & 10 & 19 & 4 & 3 & 83.42\\
2K + error corr.  & 2\,000 & std, 512    & 157/200 = 78.5\% & 21 & 14 & 8 & 0 & 83.05\\
1K + strict infer.& 1\,000 & strict, 768 & 182/200 = 91.0\% &  7 & 11 & 0 & 0 & 78.45\\
\textbf{1.2K + resid. corr.} & \textbf{1\,200} & \textbf{strict, 768}
  & \textbf{183/200 = 91.5\%} & \textbf{1} & \textbf{10} & \textbf{2} & \textbf{4} & \textbf{81.08}\\
\bottomrule
\end{tabular}
\end{table}

The best result was achieved in runs with a stricter prompt formulation and
\maxnewtokens{}. Increasing the training set to 2\,000 examples with the same
inference mode did not improve quality but reduced \texttt{full-pass@1} to 78.5\%. In contrast,
changing the inference mode without new training raised the 1K-adapter result from 82.0\% to
91.0\%. Subsequent residual correction gave a small improvement to 91.5\%, corresponding to just
one additional successfully passing test example.

Resource characteristics of the best run are presented in Table~\ref{tab2}.

\begin{table}[htbp]
\caption{Resource cost of the best pilot run.}
\label{tab2}
\centering
\begin{tabular}{lr}
\toprule
Metric & Value\\
\midrule
Training time                     & 302.3~min\\
Evaluation time (\texttt{test\_200}) & 180.9~min\\
Peak RAM during training          & 14\,582.7~MB\\
Peak RAM during evaluation        & 6\,537.1~MB\\
Mean time per token               & 169.50~ms\\
95th percentile time per token    & 191.94~ms\\
\bottomrule
\end{tabular}
\end{table}

Remaining errors in the best run concentrate primarily at L2. In particular, in some responses for
\texttt{StatefulSet} the \texttt{volumeMounts} field is placed in the wrong part of the manifest,
and in RBAC rules the required \texttt{verbs} field is sometimes absent. This shows that further
improvement requires targeted work on residual schema errors, not just increasing training data
volume.

%
% ============================================================
\section{Discussion}
% ============================================================

\subsection{Role of Strict Output Format for Small Models}

The pilot results show that in Kubernetes YAML generation, the quality of a small model depends
not only on the training corpus size but also on how strictly the output format is specified. The
main improvement was not obtained after increasing the corpus to 2K but after stabilizing the
inference mode: a strict prompt, prohibition of explanations outside YAML, and increasing
\texttt{max\_new\_tokens} to 768.

For Kubernetes YAML, the model must know typical resource templates and stably complete a
structured document without Markdown blocks and explanations, without truncating nested YAML
dictionaries and lists.

\subsection{Infrastructure Security and Privacy}

The local SLM approach is useful for infrastructure scenarios where transmitting Kubernetes
configurations to external APIs is undesirable. Such a scheme allows maintaining data control and
using the model within an isolated environment.

At the same time, the model's output cannot be considered ready for use without verification. In
the pilot, SLM is treated as a candidate generator, not as a source of trusted manifests. Each
result must pass L1--L4 validation and additional checks required in the target infrastructure.

CPU mode confirmed that the experiment can be reproduced on an ordinary laptop, but proved
expensive in time. The best run with residual correction took approximately 5~hours for training
and approximately 3~hours for evaluating \texttt{test\_200}. In practical applications, training,
experimental evaluation, and production model inference should be separated, and options for
running via OpenVINO or GPU should be checked separately.

\subsection{Representativeness and Distribution Shift}

The main risk of the pilot is the synthetic nature of the corpus and test set. Since
\texttt{test\_200} was formed from a closely related synthetic distribution, the result of
\texttt{91.5\% full-pass@1} characterizes in-domain generalization, not transfer to arbitrary
real-world production manifests.

The \texttt{real\_reverse} stream, based on real Kubernetes YAML, remains an important direction
for method development. However, in the best pilot result, it was not the primary source of
training data. The corpus was formed primarily through \texttt{synthetic\_direct} and targeted
error-correction examples.

Therefore, the external validity of the result is limited. The pilot demonstrates the pipeline's
feasibility on a fixed test set but does not substitute evaluation on an independent corpus of real
Kubernetes manifests.

\subsection{Limitations and Applicability Boundaries}

The method and pilot have the following limitations:
\begin{itemize}
  \item L3 is implemented as \texttt{L3-lite}, not as full Kubernetes semantic validation.
  \item L4 is counted on a minimal set of critical policies; other Checkov/Trivy findings are
        saved as warnings.
  \item Server-side validation via \texttt{kubectl --dry-run=server} was not performed in the
        pilot.
  \item \texttt{full-pass@5}, multiple training seeds, and Wilson confidence intervals were not
        computed.
  \item Evaluation was performed via Hugging Face Transformers and PEFT on CPU, so latency cannot be
        directly compared with OpenVINO deployment or GPU inference.
  \item The \texttt{BLEU\_auxiliary} metric is insensitive to YAML schema and semantic errors and
        was used solely for reference purposes.
\end{itemize}

For reproducibility, the following are fixed: Kubernetes schema versions, \texttt{kubeconform},
\texttt{Checkov}, \texttt{Trivy}, inference mode parameters, seed, LoRA configuration, local
schema cache, and split/validation artifacts. Corpus expansion, large-scale \texttt{real\_reverse},
pass@5, multiple training seeds, and an independent test on real YAML artifacts remain directions
for further evaluation.

%
% ============================================================
\section{Conclusion}
% ============================================================

This paper proposes the context-instrumental data distillation method for specializing small
language models in the Kubernetes domain. The main contribution is transferring formal
verification tools (\texttt{kubeconform}, \texttt{Checkov}, \texttt{Trivy}, L3-lite validator)
from the evaluation stage to the source corpus filtering stage before SLM fine-tuning.

In the pilot implementation on a laptop, the primary student model was
\qwenmodel{}, which after LoRA fine-tuning achieved
\fullpassbest{} (183/200) on the fixed \texttt{test\_200}. The key empirical
finding is that for Kubernetes YAML, the inference mode was more important than simply increasing
the training split: a strict prompt formulation and increasing the maximum number of new tokens
during generation to 768 provided a greater quality improvement than corpus expansion.

The obtained results should be treated as a pilot confirmation of pipeline feasibility, not as
evidence of quality on an independent corpus of real production manifests.

\begin{credits}
\subsubsection{\discintname}
The authors have no competing interests to declare that are relevant to the content of this
article.
\end{credits}

%
% ---- Bibliography ----
%

\end{document}